\documentclass[letterpaper, 10 pt, conference]{ieeeconf}  

\IEEEoverridecommandlockouts                              

\usepackage{graphics} 
\usepackage{epsfig} 
\usepackage{amsmath} 
\usepackage{amssymb}  
\usepackage{siunitx}
\usepackage{hyperref}
\usepackage[percent]{overpic}
\makeatletter\newcommand{\manuallabel}[2]{\def\@currentlabel{#2}\label{#1}}\makeatother
\usepackage{fancyhdr}
\fancypagestyle{plain}{%
	\fancyhf{} 
	\fancyfoot[C]{ © 2020 IEEE.  Personal use of this material is permitted.  Permission from IEEE must be obtained for all other uses, in any current or future media, including reprinting/republishing this material for advertising or promotional purposes, creating new collective works, for resale or redistribution to servers or lists, or reuse of any copyrighted component of this work in other works.}
	
}
\title{\LARGE \bf
Bounded Haptic Teleoperation of a Quadruped Robot's Foot Posture for Sensing and Manipulation
}

\author{Guiyang Xin$^1$, Joshua Smith$^1$, David Rytz$^2$, Wouter Wolfslag$^1$, Hsiu-Chin Lin$^3$ and Michael Mistry$^1$
\thanks{$^1$ G. Xin, J. Smith, W. Wolfslag, and M. Mistry are with Institute of Perception, Action and Behaviour, School of Informatics, The University of Edinburgh, 47 Potterrow, Edinburgh, UK
        {\tt\small guiyang.xin@ed.ac.uk}}%
\thanks{$^2$ D. Rytz is with the Oxford Robotics Institute at the University of Oxford, United Kingdom.}%
\thanks{$^3$ H.-C. Lin is with the School of Computer Science/Department of Electrical and Computer Engineering at McGill University, Canada.}%
}

\begin{document}

\maketitle

\thispagestyle{fancy}%
\begin{abstract}

This paper presents a control framework to teleoperate a quadruped robot's foot for operator-guided haptic exploration of the environment. Since one leg of a quadruped robot typically only has 3 actuated degrees of freedom (DoFs), the torso is employed to assist foot posture control via a hierarchical whole-body controller. The foot and torso postures are controlled by two analytical Cartesian impedance controllers cascaded by a null space projector. The contact forces acting on supporting feet are optimized by quadratic programming (QP). The foot's Cartesian impedance controller may also estimate contact forces from trajectory tracking errors, and relay the force-feedback to the operator. A 7D haptic joystick, Sigma.7, transmits motion commands to the quadruped robot ANYmal, and renders the force feedback. Furthermore, the joystick's motion is bounded by mapping the foot's feasible force polytope constrained by the friction cones and torque limits in order to prevent the operator from driving the robot to slipping or falling over. Experimental results demonstrate the efficiency of the proposed framework.

\end{abstract}

\section{INTRODUCTION}

Haptic teleoperation plays an important role in robotic application cases, such
as robot-assisted surgery and nuclear facility decommissioning. Vision-based
teleoperation can be enhanced by haptic rendering., e.g., an user might want to remotely inspect a specific area through touching.
Haptic feedback also helps visual perception when vision systems lose efficacy in some special cases,
e.g., lidar can give wrong terrain maps when the terrain is covered by liquid or grass. 
Exploring the environment using tactile sensors could mitigate errors in
vision system results. In the literature a number of haptic teleoperation works
with single arm robots can be found, whereas teleoperating quadruped robots to
do manipulation and exploration has been studied little.

Researchers have been working on taking full advantage of multi-legged robots
to be more versatile. For humanoid robots it is natural to do manipulation
using arms \cite{jenkins2006uncovering}\cite{atkeson2015no}. Equipping
quadruped robots with an additional arm that is dedicated to manipulation tasks
is a common way to enhance locomanipulation
\cite{rehman2016towards}\cite{spotmini}\cite{bellicoso2019alma} due to the fact that the leg
of a quadruped usually does not have enough DoFs nor a sufficiently large
workspace to be competent at six dimensional manipulation in Cartesian space.
Whole-body controllers can implement foot posture control by coordinating the leg and base, which greatly extends the versatility of quadruped robots in real world deployment.

A fully optimization-based whole-body controller called hierarchical quadratic programming (HQP) has been widely used on legged robots \cite{de2010feature}\cite{escande2014hierarchical}\cite{Herzog2016}\cite{DarioBellicoso2016}.
Multiple tasks are encoded by sequential strictly null-space prioritized
QPs, which solve for the torque commands while taking into account joint space dynamics and physical constraints.
Solving several QPs online is still time consuming, although decomposition
methods are developed to reduce the number of decision variables.
Usually three QPs have to be solved for dynamic feasibility, operational tasks and saving energy \cite{bellicoso2019alma}.
Alternatively a non-strictly prioritized weighted QP (WQP) is used to avoid solving sequential QPs by stacking task equations into a weighted quadratic cost function \cite{de2009prioritized}\cite{Feng2015}.

Alternatively, projected inverse dynamics is proposed by \cite{Aghili2005}\cite{Aghili2016}\cite{lin2018projected} to split up motion space and constraint space so that only one QP in constraint space needs to be solved. In \cite{Xin2018}, we derived operational space dynamics for floating base robots and then obtained analytical Cartesian impedance controllers in motion space. Although a QP optimization is still required to solve inequality constraints, the analytical Cartesian impedance controller gives us the ability to set gains based on operational space inertia \cite{Angelini2019} and to estimate external forces because of properly scaled gains. In this paper, we will apply that analytical method to haptic teleoperation for sensing and manipulation. 

Since the hierarchical controller will sacrifice low level tasks to satisfy high level tasks, one may not be able to fulfill the constraints at the lower level. For example, if the foot controller has higher priority than the base controller, the base may move outside the support polygon and become unstable.
To avoid such failures, we need to set boundaries to restrict teleoperation movement.
The commanded end-effector forces should also be bounded based-on physical criteria, as high acting force may cause the supporting feet to slip and/or
lose balance due to moment around support polygon edge generated by the acting force.
Papers about motion planning have discussed balance maintenance for multiple
contact cases \cite{harada2006dynamics}\cite{dai2014whole}.
This paper will present how to appropriately restrict operator commands by mapping the
end-effector feasible wrench polytope (FWP)
\cite{ferrolho2019comparing}\cite{orsolino2018application} to joystick boundaries. 

The contribution of this paper is twofold. Firstly, we extend our hierarchical Cartesian impedance controller to adapt to foot posture teleoperation, and the impedance controller is experimentally proved to be accurate enough to estimate contact forces.
Secondly, the range of the teleoperation joystick is bounded with respect to the physical constraints to ensure safe teleoperation. To the best of authors' knowledge, this is the first time to achieve haptic teleoperation of foot posture for quadruped robots in the literature.


\section{Control framework}

\begin{figure*}[ht]
\centering
 \includegraphics[width=0.95\linewidth]{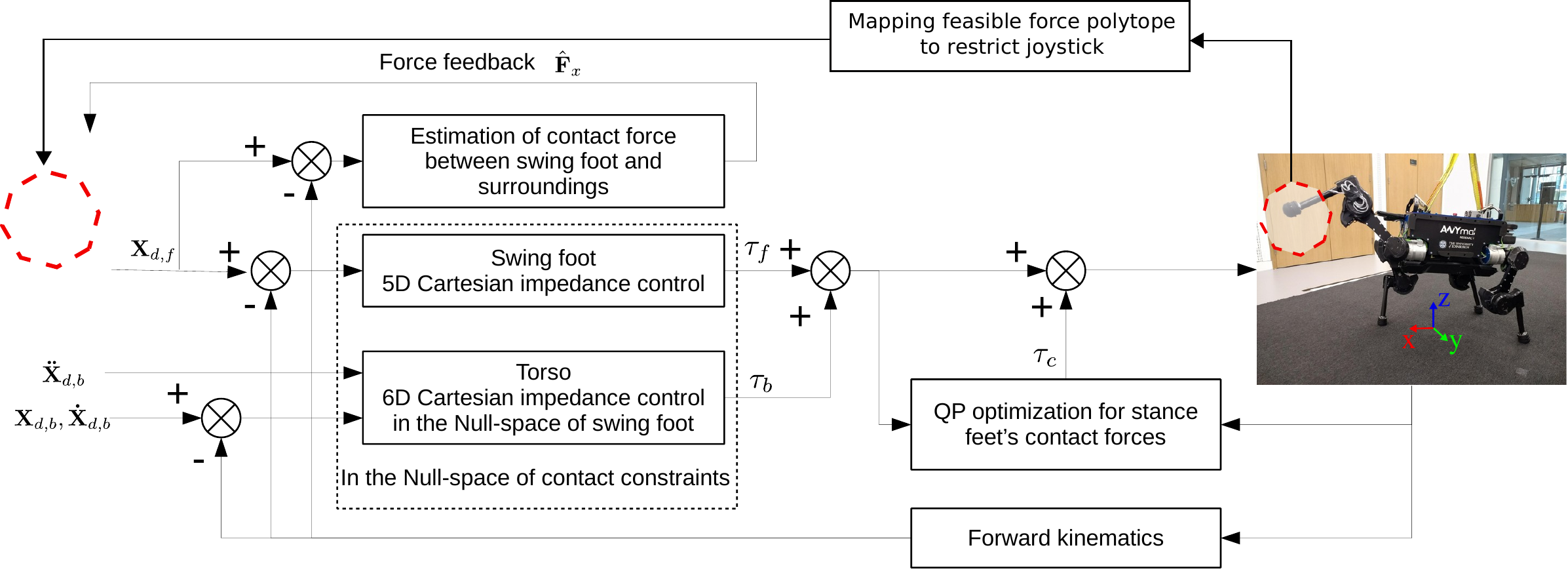}
 \caption{Schematic of foot posture controller with haptic teleoperation.}
 \label{f:framework}
\end{figure*}

The joint space dynamics equation of a quadruped robot is
\begin{equation}\label{dynamics}
    \mathbf{M}\mathbf{\ddot{q}}+\mathbf{h}=\mathbf{B}\boldsymbol{\tau}+\mathbf{J}_c^{\top}\boldsymbol{\lambda}_c
\end{equation}
where $\mathbf{q}=\begin{bmatrix}\mathbf{x}_b^\top &\mathbf{q}_j^\top\end{bmatrix}^\top$ denotes the vector of $n$  actuated joint positions ($\mathbf{q}_j \in \mathbb{R}^n$) and floating base position and orientation ($\mathbf{x}_b \in
    SE(3)$), $\mathbf{M} \in \mathbb{R}^{(n+6)\times(n+6)}$ is the inertia matrix, $\mathbf{h} \in \mathbb{R}^{n+6}$ is the generalized vector containing Coriolis, centrifugal and gravitational effects,
$\boldsymbol{\tau} \in \mathbb{R}^{n+6}$ is the vector of torques, $\mathbf{J}_c \in \mathbb{R}^{3k\times(n+6)}$ is the constraint Jacobian that describes $3k$ linearly independent constraints, $k$ denotes the number of supporting feet,
$\boldsymbol{\lambda}_c \in \mathbb{R}^{3k}$ are constraint forces acting on supporting feet, and
\begin{equation}\label{B}
    \mathbf{B}=\begin{bmatrix}
        \mathbf{0}_6 & \mathbf{0}   \\
        \mathbf{0}   & \mathbf{I}_n
    \end{bmatrix}
\end{equation}
is the selection matrix with $n$ dimensional identity matrix $\mathbf{I}_n$. 
Eq. \eqref{dynamics} is subject to the following physical constraints,
\begin{equation}\label{e:no-slipping}
    \mathbf{J}_c\ddot{\mathbf{q}}+\dot{\mathbf{J}}_c\dot{\mathbf{q}}=\mathbf{0}
\end{equation}
\begin{equation}\label{e:torque_limitation}
    \tau_{min}\leq\tau\leq\tau_{max}, \quad i=1,\dots,n
\end{equation}
\begin{equation}\label{e:friction_cone}
    \mu\lambda_{z,i}\geq\sqrt{\lambda_{x,i}^2+\lambda_{y,i}^2}, \quad i=1,\dots,k
\end{equation}
\begin{equation}\label{e:unilateral}
    \lambda_{z,i}\geq0, \quad i=1,\dots,k
\end{equation}
where Eq. \eqref{e:no-slipping} means supporting feet should not move relative to the ground, and Eqs. \eqref{e:torque_limitation}, \eqref{e:friction_cone} and \eqref{e:unilateral} represent torque saturation, friction cone and unilateral constraints respectively.

\subsection{Cartesian impedance controller}

In paper \cite{mistry2012operational}, the authors suggested to use a null-space projector $\mathbf{P}=\mathbf{I}-\mathbf{J}_c^{+}\mathbf{J}_c$ to split up \eqref{dynamics} into two orthogonal subspaces,
\begin{equation}\label{e:p_space}
    \mathbf{P}\mathbf{M}\mathbf{\ddot{q}}+\mathbf{P}\mathbf{h}=\mathbf{P}\mathbf{B}\boldsymbol{\tau}=\boldsymbol{\tau}_m
\end{equation}
\begin{equation}\label{e:i_p_space}
    \mathbf{(I-P)}(\mathbf{M}\mathbf{\ddot{q}}+\mathbf{h})=\mathbf{(I-P)}\mathbf{B}\boldsymbol{\tau}+\mathbf{J}_c^{\top}\boldsymbol{\lambda}_c=\boldsymbol{\tau}_c+\mathbf{J}_c^{\top}\boldsymbol{\lambda}_c
\end{equation}
Eq. \eqref{e:p_space} represents motion space that is not affected by contact forces of supporting feet; meanwhile, Eq. \eqref{e:i_p_space} represents dynamics in constraint space. The advantage of this separation is that we can derive an analytical Cartesian impedance controller with Eq. \eqref{e:p_space} and solve a QP optimization to generate an extra torque command satisfying constraints of Eqs. (4-6) with Eq. \eqref{e:i_p_space} as shown in subsection II-B. The final torque command is composed of two parts as follows,
\begin{equation}\label{e:tau}
    \boldsymbol{\tau}=(\mathbf{PB})^{+}\mathbf{P}(\mathbf{J}_s^{\top}\mathbf{F}_s+\mathbf{N}_s\mathbf{J}_b^{\top}\mathbf{F}_b)+\boldsymbol{\tau}_c
\end{equation}
where the first right side term is equivalent to $\boldsymbol{\tau}_m$ resulting from Cartesian impedance control and $\boldsymbol{\tau}_c$ is the result of QP optimization. $\mathbf{J}_s$ denotes the Jacobian matrix of a swing foot, the dimension of which depends on the control task. Since we want to control both the position and orientation of the foot, $\mathbf{J}_s \in \mathbb{R}^{6\times(n+6)}$. $\mathbf{F}_s$ and $\mathbf{F}_b$, derived from operational space dynamics, are Cartesian impedance controllers for the swing foot and the base. $\mathbf{N}_s$ is the dynamic consistent null-space projector \cite{mistry2012operational} of the swing foot, which enforces strictly hierarchical priorities. In the case of foot posture control, $\mathbf{N}_s$ will deal with the overlap between $\mathbf{J}_s$ and $\mathbf{J}_b$, leading to the convenience of leaving base Jacobian $\mathbf{J}_b$ to be always a $\mathbb{R}^{6\times(n+6)}$ matrix. As the torso is in the null-space of the swing foot, the torso is enforced to satisfy the swing foot's motion requirement, which results in automatic motion coordination and reachability extension of foot.

The generic equation of the Cartesian impedance control law for $\mathbf{F}_s$ and $\mathbf{F}_b$ is
\begin{equation}\label{e:control_law}
    \mathbf{F}_i=\mathbf{h}_{c,i}+\boldsymbol{\Lambda}_{c,i}\mathbf{\ddot{x}}_{d,i}-\mathbf{D}_{d,i}\mathbf{\dot{e}}_{i}-\mathbf{K}_{d,i}\mathbf{e}_i, \quad i=s \text{ or }b
\end{equation}
where $\mathbf{h}_c$ represents the operational space gravity compensation vector, $\boldsymbol{\Lambda}_c$ denotes the operational space inertia matrix, $\mathbf{D}_d$ and $\mathbf{K}_d$ are diagonal matrices containing desired damping and stiffness gains, $\mathbf{e}=\mathbf{x}-\mathbf{x}_d$ is the pose error of end-effector (either the base or the swing foot), $\mathbf{\ddot{x}}_d$ denotes desired acceleration of end-effector. We refer to \cite{Xin2018} for more details of $\mathbf{h}_c$ and $\boldsymbol{\Lambda}_c$.

The impedance control law Eq. \eqref{e:control_law} leads to the following impedance behavior under external disturbances
\begin{equation}\label{impedance}
    \boldsymbol{\Lambda}_c\mathbf{\ddot{e}}+\mathbf{D}_d\mathbf{\dot{e}}+\mathbf{K}_d\mathbf{e}=\mathbf{F}_x
\end{equation}
where $\mathbf{F}_x=\begin{matrix}[\mathbf{f}_x^\top &\mathbf{m}_x^\top]\end{matrix}^\top \in \mathbb{R}^6$ is the external wrench acting on the end-effector. Obviously if there is no external disturbance, the robot will track desired trajectory accurately with the assumption of using a perfect model. Conversely, we could measure motion error and then estimate external disturbances~\cite{lin2018projected},

\begin{equation}\label{e:estimation}
    \hat{\mathbf{F}}_x=\boldsymbol{\Lambda}_c\mathbf{\ddot{e}}+\mathbf{D}_d\mathbf{\dot{e}}+\mathbf{K}_d\mathbf{e}
\end{equation}

Model error always exists for real robot platforms, leading to a
$\hat{\mathbf{F}}_x$ caused by both model error and disturbances. If the model
error is much smaller than disturbances, $\hat{\mathbf{F}}_x$ is accurate enough to
be a contact force estimator. Here in our experiments, we employ this
estimation as haptic feedback for teleoperation, and thus do not require a force/torque sensor at the point of contact.

Note that the torso motion error doesn't affect the foot position error because the
torso is controlled in the null-space of the manipulation foot.
If the base were fixed, the same external force, $\mathbf{F}_x$, acting on the
foot would result in the same motion error, $\mathbf{e}_s$, as when running our
hierarchical whole-body controller with a floating base.
The difference only exists in joint errors. We can always only
measure the motion of the foot and then use Eq. \eqref{e:estimation} to estimate
external force $\mathbf{F}_x$ without any torso motion error interfering.

\begin{equation}\label{e:estimation}
    \hat{\mathbf{F}}_x=\boldsymbol{\Lambda}_{c,s}\mathbf{\ddot{e}}_s+\mathbf{D}_{d,s}\mathbf{\dot{e}}_s+\mathbf{K}_{d,s}\mathbf{e}_s
\end{equation}


\subsection{QP optimization}

The control structure is depicted in Fig. \ref{f:framework} where the inputs of QP optimization in constraint space are torques for the motion and state feedback, which implies the motion can affect contact forces $\boldsymbol{\lambda}_c$ as $\mathbf{\ddot{q}}$ are involved in Eq. \eqref{e:i_p_space}. We solve the forward dynamics of Eq. \eqref{e:p_space} to derive the equation of $\mathbf{\ddot{q}}$ with respect to $\boldsymbol{\tau}$, and then substitute it into Eq. \eqref{e:i_p_space} to eliminate $\mathbf{\ddot{q}}$. Since $\mathbf{PM}$ is not invertible, we resort to a trick (provided in \cite{mistry2012operational}\cite{Aghili2005}) of using $\mathbf{M}_c=\mathbf{PM}+\mathbf{I}-\mathbf{P}$ to replace $\mathbf{PM}$ in Eq. \eqref{e:p_space} as $(\mathbf{I}-\mathbf{P})\mathbf{\ddot{q}}=\mathbf{\dot{P}}\mathbf{\dot{q}}$ holds, leading to
\begin{equation}\label{e:qddot}
    \mathbf{\ddot{q}}=\mathbf{M}_c^{-1}(\boldsymbol{\tau}_m-\mathbf{Ph}+\mathbf{\dot{P}}\mathbf{\dot{q}})
\end{equation}
where $\boldsymbol{\tau}_m$ is the result from Cartesian impedance control in our case. Subsequently, substituting \eqref{e:qddot} into Eq. \eqref{e:i_p_space} yields
\begin{equation}\label{e:lambda}
    \boldsymbol{\lambda}_c=(\mathbf{J}_c)^{+}\big[(\mathbf{I}-\mathbf{P})[\mathbf{M}\mathbf{M}_c^{-1}(\boldsymbol{\tau}_m-\mathbf{Ph}+\mathbf{\dot{P}}\mathbf{\dot{q}})+\mathbf{h}]-\boldsymbol{\tau}_c\big]
\end{equation}
Therefore, the QP in constraint space with respect to $\boldsymbol{\tau}_c$ can be defined as
\begin{equation}\label{e:qp}
    \begin{aligned}
        & \underset{\boldsymbol{\tau}_c}{\text{minimize}}
        & & \frac{1}{2}\|\boldsymbol{\tau}_c\|_2^2 \\
        & \text{subject to}
        & &  \boldsymbol{\tau}_{\text{min}}-\boldsymbol{\tau}_m\leq\boldsymbol{\tau}_c\leq\boldsymbol{\tau}_{\text{max}}-\boldsymbol{\tau}_m\\
        &&& \mu(\boldsymbol{\eta}-\mathbf{J}_c^{+}\boldsymbol{\tau}_c)_{z,i}\geq\sqrt{\lambda_{x,i}^2+\lambda_{y,i}^2} \\
        &&& (\boldsymbol{\eta}-\mathbf{J}_c^{+}\boldsymbol{\tau}_c)_{z,i} \geq 0
    \end{aligned} 
\end{equation}
where $\boldsymbol{\eta}=(\mathbf{J}_c)^{+}\big(\mathbf{I}-\mathbf{P})[\mathbf{M}\mathbf{M}_c^{-1}(\boldsymbol{\tau}_m-\mathbf{Ph}+\mathbf{\dot{P}}\mathbf{\dot{q}})+\mathbf{h}]$ abbreviates the first right side term of Eq. \eqref{e:lambda}. $\boldsymbol{\tau}_m$ in Eq. \eqref{e:qp} is regarded as a constant vector. The QP may have no solution if the desired trajectory is not physically feasible, which can be solved by using a motion planner subject to physical constraints. In the case of teleoperation, this motion is generated by the operator. Particularly when the robot has interaction during a manipulation mission, the operator may push the joystick too much resulting in slipping. The next section will discuss how to set boundaries on the  joystick in order to avoid the operator breaking physical constraints.


\section{Feasibility boundaries for teleoperation}\label{s:feasibility_boundaries}
An user may operate the joystick to execute commands that are physically incapable for the robots. 
Instead of simply ignoring such infeasible commands, we propose to set motion boundaries on the joystick in order to give operators early notifications of approaching dynamically infeasible areas.

\subsection{Boundaries on CoM position}

To limit CoM motion, we look for the maximum range in
each direction from the current foot position that will keep the CoM within the support
polygon. Theoretically, we can use 6 DoFs to control foot posture meanwhile
using the remaining 3 DoFs to control the torso.
The null-space projector will automatically use the 3 DoFs for the torso to guarantee the foot trajectory tracking if the 6 DoFs assigned to foot are not enough to achieve desired motion.
Solving forward dynamics and simulating the system for a time horizon can give us predicted robot state.
If we want to calculate the boundary in Cartesian space, that will be time consuming to predict.
Instead of directly finding the boundaries of the foot, we shrink the supporting polygon to be
the boundary of CoM.
When the CoM is close to the boundary, joystick stiffness will increase to stop joystick moving along that direction.

\subsection{Boundaries for feasible acting forces} 

Since the joystick sends position commands to the impedance controller, the impedance controller generates acting forces based on the position errors. We can restrict the workspace of joystick to saturate the acting force. Firstly, we need to compute the set of forces at the end-effector respecting joint torque limits and the contact force constraints. Such a set has been used to enhance motion planning of robot manipulators~\cite{ferrolho2019comparing}. The operational-space impedance controller gains from \eqref{impedance} are used to transform the set of feasible end-effector forces to feasible displacements, and are subsequently mapped to feasible joystick commands. The joystick will run a varying impedance controller to sharply increase the stiffness when the operator is moving the joystick into a infeasible position.

To compute the set of feasible forces at the end-effector, we extend the computational framework from \cite{orsolino2018application}, which computes the set of feasible wrenches applied at the torso of the robot.
As in that framework, we note the structure of the equations of motion \eqref{dynamics}:
\begin{equation}
\resizebox{\columnwidth}{!}{$
   \begin{bmatrix} \mathbf{d}_{b} \\ \mathbf{d}_{0} \\ \mathbf{d}_{1} \\ \mathbf{d}_{2} \\ \mathbf{d}_{3}
   \end{bmatrix} = 
    \begin{bmatrix}\mathbf{0} & \mathbf{0} & \mathbf{0} & \mathbf{0} & \mathbf{0} \\ \mathbf{0} & \mathbf{I} & \mathbf{0} & \mathbf{0} & \mathbf{0} \\
                   \mathbf{0} & \mathbf{0} & \mathbf{I} & \mathbf{0} & \mathbf{0} \\ \mathbf{0} & \mathbf{0} & \mathbf{0} & \mathbf{I} & \mathbf{0} \\
                   \mathbf{0} & \mathbf{0} & \mathbf{0} & \mathbf{0} & \mathbf{I}\end{bmatrix}
    \begin{bmatrix} \mathbf{0} \\ \boldsymbol{\tau}_{0} \\ \boldsymbol{\tau}_{1} \\ \boldsymbol{\tau}_{2} \\ \boldsymbol{\tau}_{3}
    \end{bmatrix} +
    \begin{bmatrix}
    \mathbf{J}_{0,\text{b}}^{\top} & \mathbf{J}_{1,\text{b}}^{\top} & \mathbf{J}_{2,\text{b}}^{\top} & \mathbf{J}_{3,\text{b}}^{\top} \\
    \mathbf{J}_{0,0}^{\top} & \mathbf{0} & \mathbf{0} & \mathbf{0} \\
    \mathbf{0} & \mathbf{J}_{1,1}^{\top} & \mathbf{0} & \mathbf{0} \\
    \mathbf{0} & \mathbf{0} & \mathbf{J}_{2,2}^{\top} & \mathbf{0} \\
    \mathbf{0} & \mathbf{0} & \mathbf{0} & \mathbf{J}_{3,3}^{\top}
    \end{bmatrix}
    \begin{bmatrix} \boldsymbol{\lambda}_0 \\ \boldsymbol{\lambda}_1 \\ \boldsymbol{\lambda}_{2} \\ \boldsymbol{\lambda}_{3}
    \end{bmatrix}$
} 
\label{eq:eom_structured}
\end{equation}
where we have split the inertial forces ($\mathbf{d} =  \mathbf{M}\mathbf{\ddot{q}} + \mathbf{h}$), the joint-torques, and the contact forces into parts associated with the base and the separate legs and feet. 
Leg number 0 performs the manipulation, meaning the associated $\lambda_0$ will be 0.

The bottom four rows provide a constraint between joint torques and contact forces for each leg:
\begin{equation}
    \boldsymbol{\lambda}_i = \mathbf{J}_{i,i}^{-\top}(\mathbf{d}_i - \boldsymbol{\tau}_i), \quad  i = 0,1,2,3
    \label{torque_feasibility}
\end{equation}
These equality constraints map the hypercube of allowed joint torques to a polytope of contact forces. 
For the stance-legs these polytopes are intersected with the friction cone. 
Any constraints on the contact force in manipulation, for instance when interacting with fragile environments, can be added at this stage as well. 
This results in a polytope ($\mathit{FCP}_i$) of feasible contact forces for each separate leg.

Finally the top row of \eqref{eq:eom_structured} combines these polytopes with the dynamics of the robot torso, to form the feasible force polytope $\mathit{FFP}$ for the manipulating leg:
\begin{equation}
    \mathit{FFP} = \{\mathbf{f} \in \mathit{FCP}_0 \enskip | \enskip \mathbf{J}_{0,\text{b}}^{\top}\mathbf{f} \in (\mathbf{d}_\text{b} - \sum_{i=1}^3 \mathbf{J}_{i,\text{b}}^{\top}\mathit{FCP}_i)\}
    \label{eq:ffp}
\end{equation}
where a sum of polytopes is taken to be their Minkowski sum.
The polytope-operations are handled using the CDDLib\footnote{See https://github.com/JuliaPolyhedra/CDDLib.jl} library. Unfortunately the various transformations require multiple conversions between vertex and halfspace representations, making the computation of the feasible force polytope expensive. As a result the polytopes are computed offline with a simple look-up used for online access.

\begin{figure}
    \centering
    \includegraphics[width=.95\linewidth]{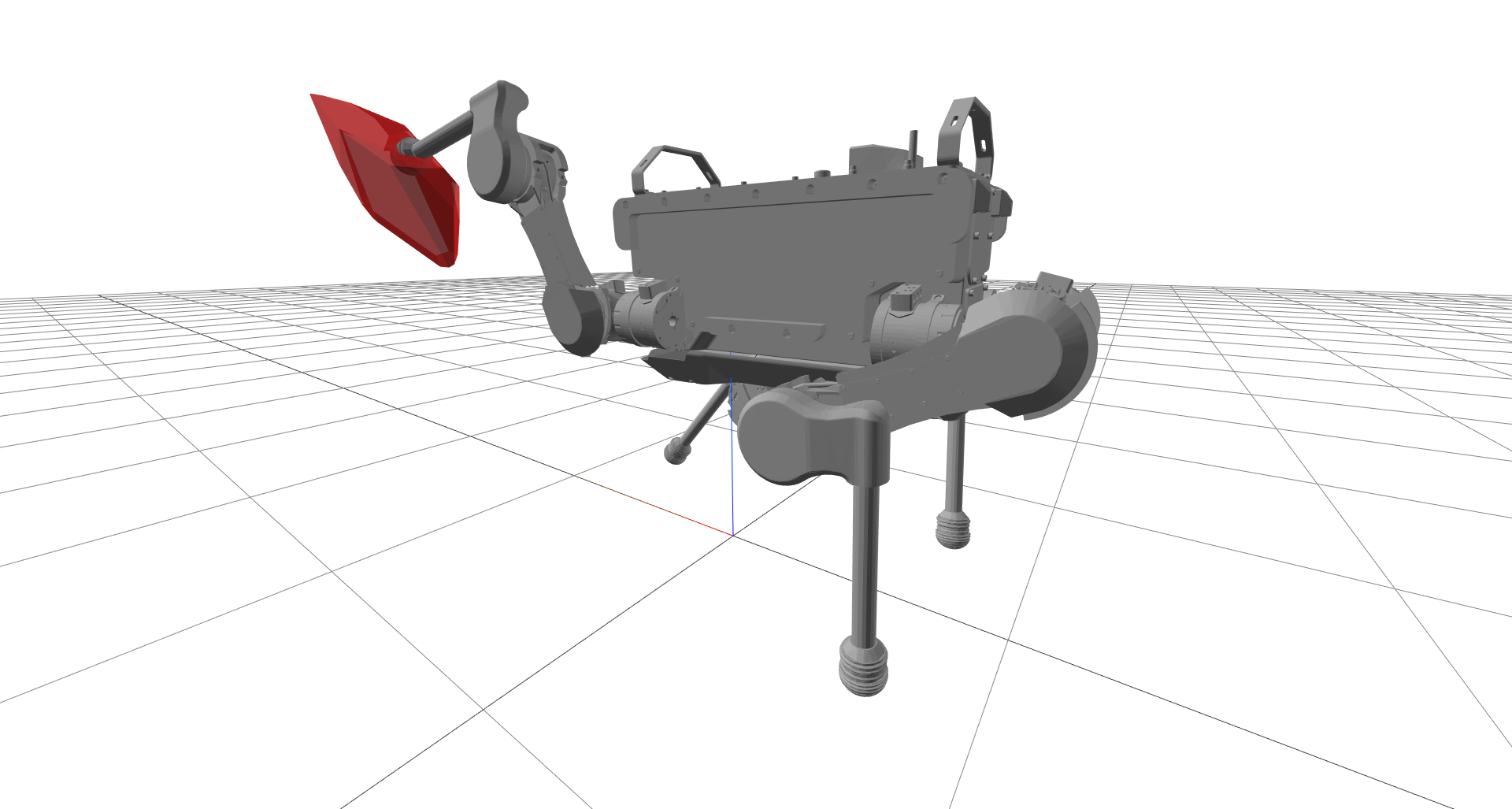}
    \caption{The feasible force polytope during the experiment. Shown in red are the forces that can be applied to the environment. The force polytope is shown with the origin shifted to the centre of the foot and using a scaling of $\SI{500}{\newton\per\metre}$.}
    \label{fig:feasible_force_polytope}
\end{figure}

Figure \ref{fig:feasible_force_polytope} shows the feasible force polytope for a configuration in simulation. In this configuration the robot can exert little force upwards, as larger forces would shift the centre of pressure beyond the support polygon. The posture of each surface of the polytope in world frame is defined by the normal vector and the maximum feasible force along this direction. In theory we need to project the contact force on all these directions (37 for this configuration) and then compare the projection with the maximum feasible force along these directions. If the projected force is greater than the maximum feasible force, that means the contact force will break physical constraints. In practice we will shrink the $\mathit{FFP}$ corresponding to a certain configuration to be the boundary on the joystick. When any direction is close to the boundary we will generate a force to prevent the joystick from entering the area and to inform the operator. 

\section{Experiments}
The proposed algorithms are implemented on the quadruped robot ANYmal\footnote{See http://www.anybotics.com/} and a Sigma.7\footnote{See http://www.forcedimension.com/} haptic joystick.
In practical manipulation cases the yaw of the foot with respect to foot frame
($z$ axis of foot frame coincides with the central axis of shank) has little to
no effect on the task, so we chose not to control it.
The remaining 5 dimensions are controlled by the foot impedance controller as shown in Fig. \ref{f:framework}. 
Not controlling the yaw helps in giving the torso controller more authority in the null-space, otherwise the torso is quite sensitive to disturbance and non-smooth motion commands.
The experiment videos can be found at: \url{https://www.youtube.com/watch?v=htI8202vfec}. 

\subsection{Haptic joystick setup}
The Sigma.7 haptic device used in Fig. \ref{f:framework} is connected to the quadruped using a WiFi connection and custom ROS node running on a different computer than the ANYmal with Ubuntu 18.04 LTS. The sensing and control cycle of Sigma.7 is \SI{100}{\hertz}, meanwhile the control loop of locomotion controller on ANYmal's PC is \SI{400}{\hertz}. The commercial version from Force Dimension comes with a SDK that allows to easily access the device pose, twist and wrench of the device end-effector while also applying a desired wrench. For this work the device pose is read and added to the current quadruped end-effector pose to generate the target pose. As the workspace of the haptic device is significantly smaller than the ANYmal one, the Sigma.7 gripper is used to reset the end-effector command. To reduce the mental load on the operator and increase the stability of the quadruped the authors implemented the following measures:
\begin{itemize}
    \item The haptic device stiffens when the quadruped is reaching feasible limits as described in section \ref{s:feasibility_boundaries}.
    \item As the quadruped might experience high contact forces the reported contact forces are only applied when the operator presses the device gripper. In case of an open gripper state the Sigma.7 device goes into it's initial base pose in the center of the workspace.
\end{itemize}

\subsection{Accuracy verification of estimated contact force}

\begin{figure}[t!]
\centering
\includegraphics [width=.48\linewidth]{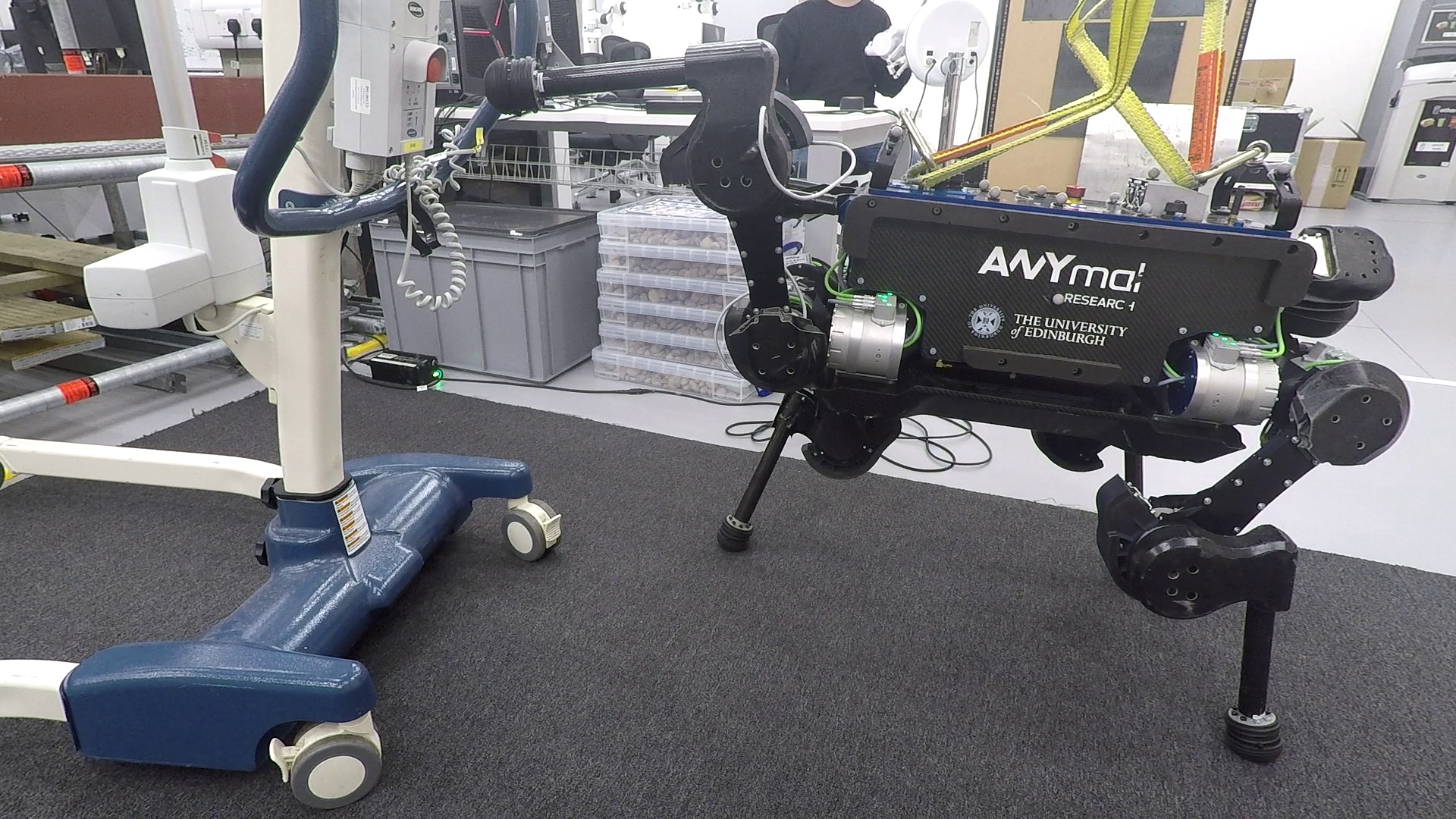}~
\includegraphics [width=.48\linewidth]{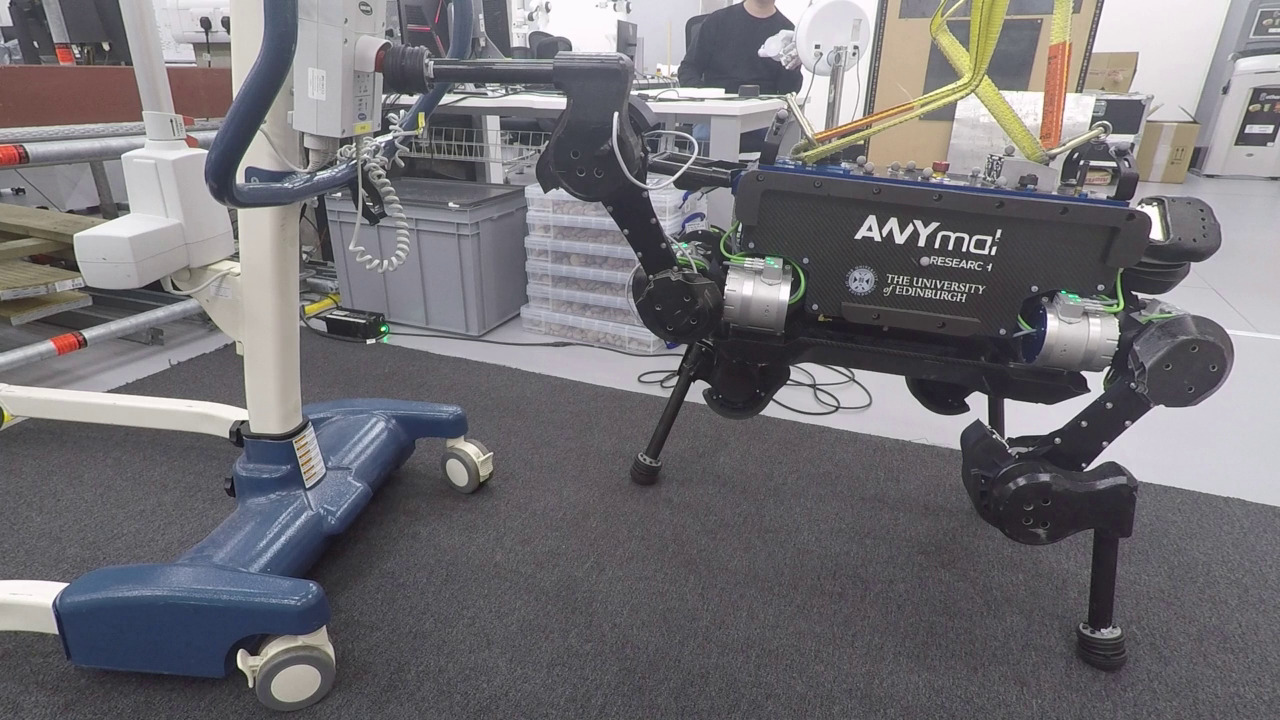}
\\ \vspace{-1mm} 
\caption{Teleoperating ANYmal to press the e-stop button of a crane.}  
\label{f:accuracy} 
\end{figure}

The Cartesian impedance controller can estimate the contact force as well as the contact wrench.
Usually we do not need wrench feedback since the foot shape is a small sphere, designed as a point foot.
The first experiment is to verify how accurate the force estimation based on impedance controller is. We teleoperated the robot to press an e-stop button of a crane as shown in \ref{f:accuracy}. The force sensor inside the foot is enabled to record sensory contact force to compare with estimated force by impedance controller. The Eq. \eqref{e:estimation} is used to compute estimated forces. However, acceleration and velocity are noisy. If the estimated force contains high frequency components it will cause oscillation on joystick when rendering the feedback forces. In the end, we used only the position error to estimate the contact forces.

\begin{figure}
    \centering
    \includegraphics [width=.95\linewidth]{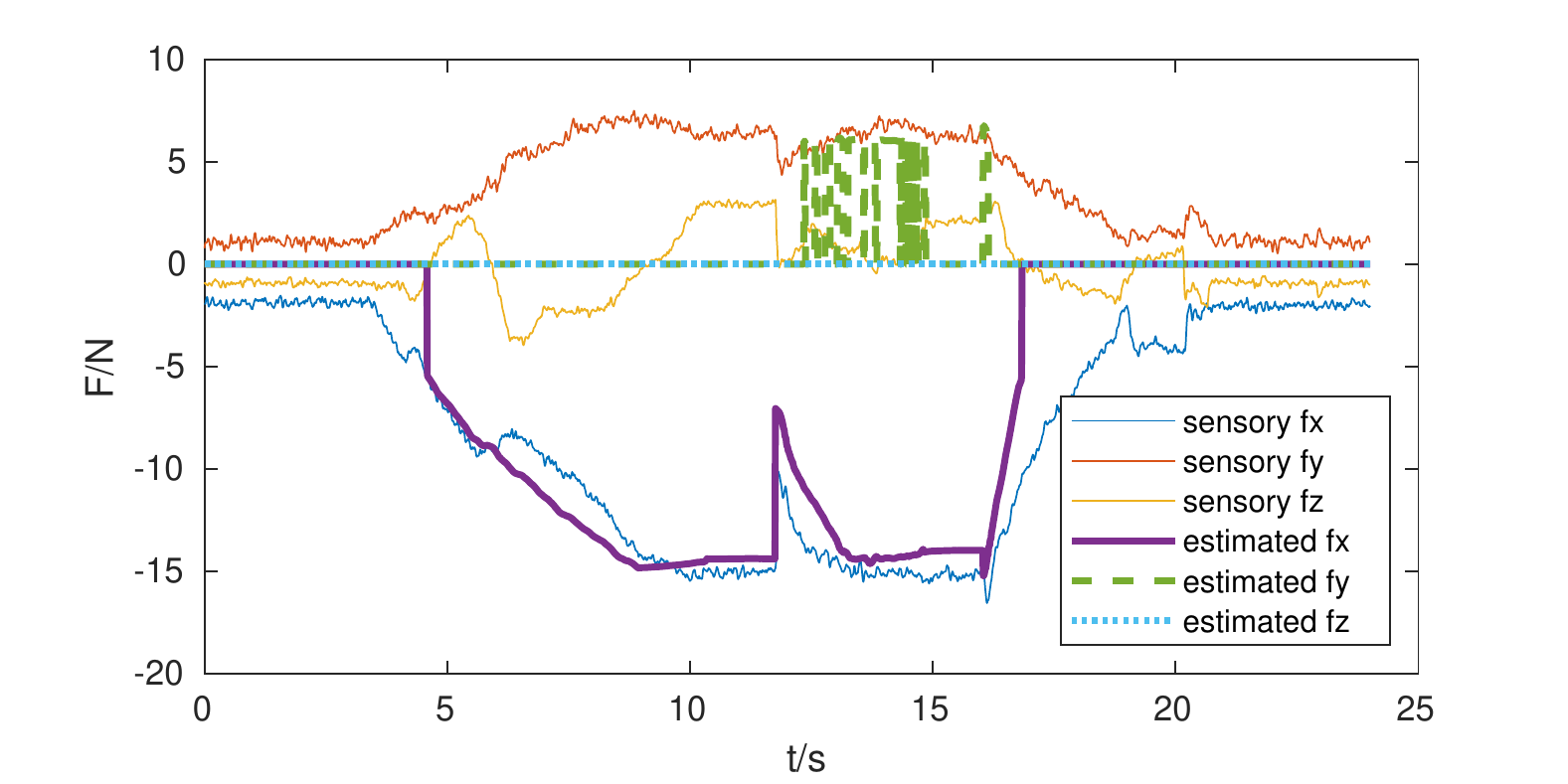}
    \\ \vspace{-1mm} 
    \caption{Force comparison between sensory results and estimated results during pressing button experiment. The button was pressed at around \SI{5}{\second}. The contact forces suddenly changed at around \SI{12}{\second} because the button started to move. After pressing down, contact forces increased again. The operator released the button at around \SI{16}{\second}. Compared to noisy sensory forces, the estimated forces are better to be used as haptic feedback.}
    \label{f:forceCompare}
\end{figure}

The recorded data is shown in Fig. \ref{f:forceCompare}. It should be noted
that we set a threshold of \SI{5}{N} to filter the estimated force, intending to discriminate non-contact error when swinging in the air.
This threshold can explain the spikes of estimated $f_y$, and also leads to
$f_z=0$ because estimated $f_z$ is always less than \SI{5}{\newton} in this
experiment. The force dropping at around the \SI{12}{\second} mark, is due to
the button being fully pressed.
When the button reaches the bottom, the contact force increases until the
operator feels that the button was completely pressed down. 

Based on Fig. \ref{f:forceCompare} the estimated force is accurate enough to be used as haptic feedback for teleoperation. Additionally compared to noisy sensory forces (the sensory result shown in Fig. \ref{f:forceCompare} has already been filtered by a second order low pass filter) the estimated force is easier to be projected onto the joystick.

{}

\subsection{Pipe manipulation}

\begin{figure}
\centering
\begin{overpic} [width=.48\linewidth]{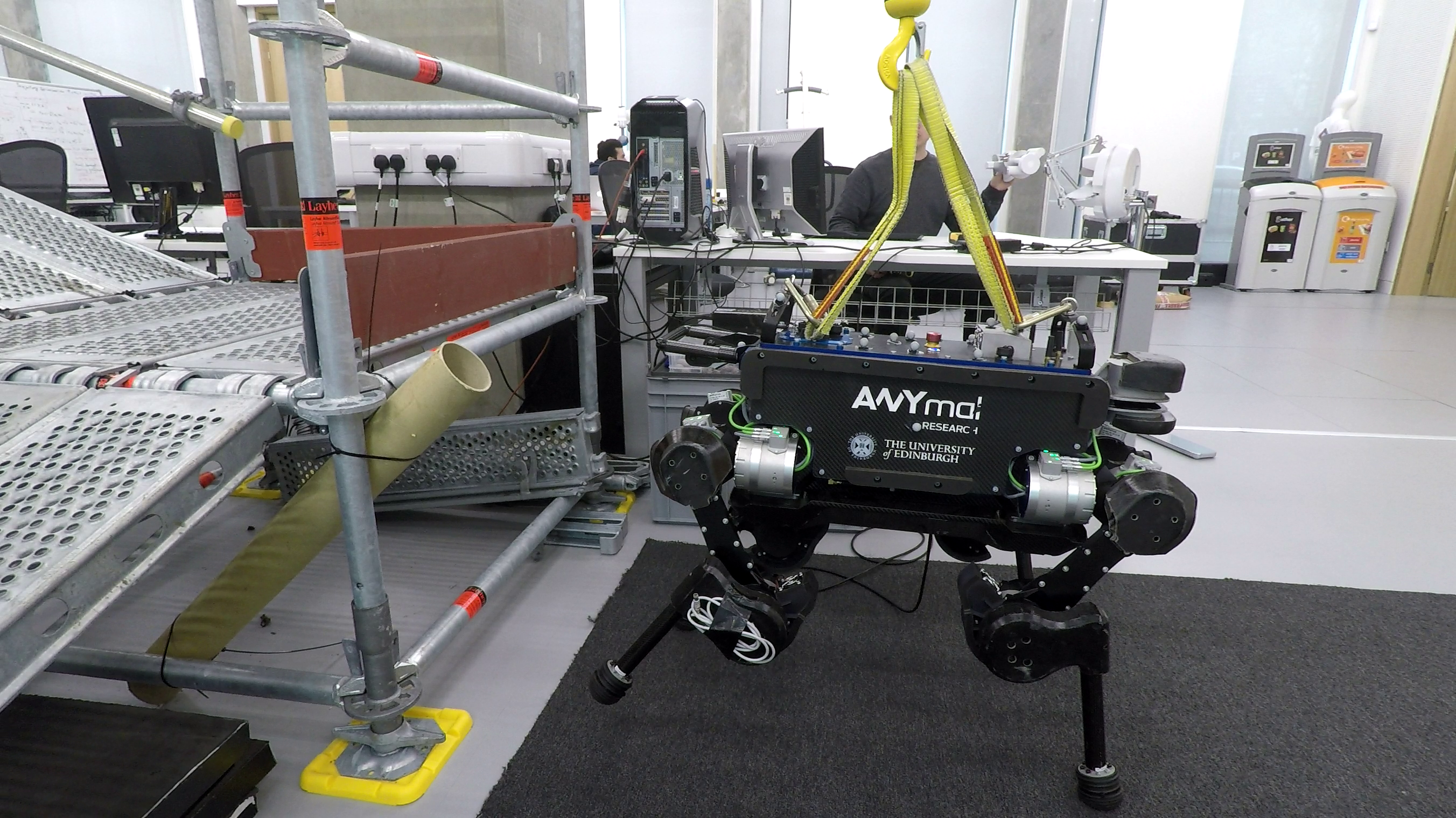}\put(90,50){\manuallabel{f:pipInsert1}{(a)}\ref{f:pipInsert1}}\end{overpic}~
\begin{overpic} [width=.48\linewidth]{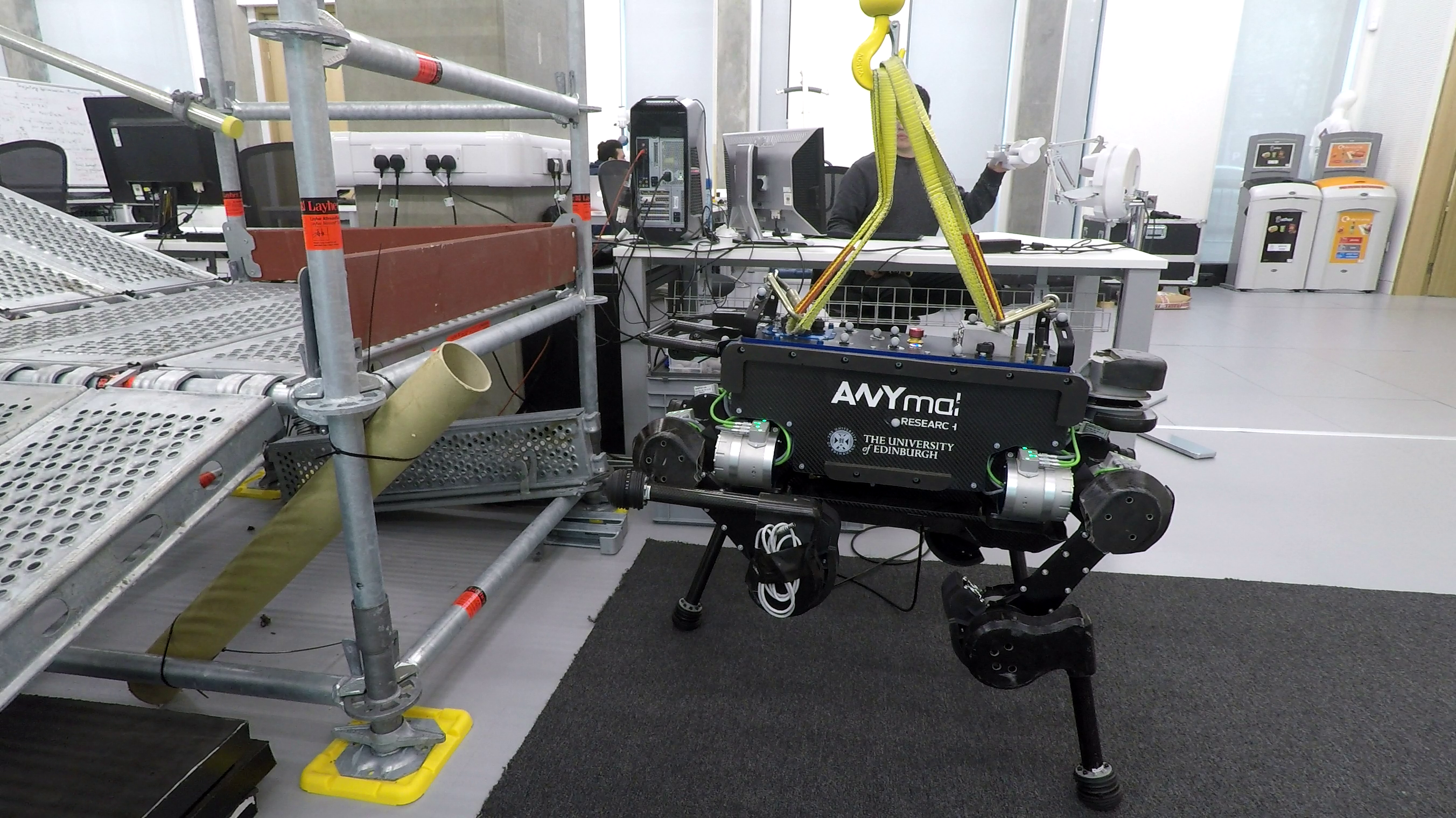}\put(90,50){\manuallabel{f:pipInsert2}{(b)}\ref{f:pipInsert2}} \end{overpic}\\
\begin{overpic} [width=.48\linewidth]{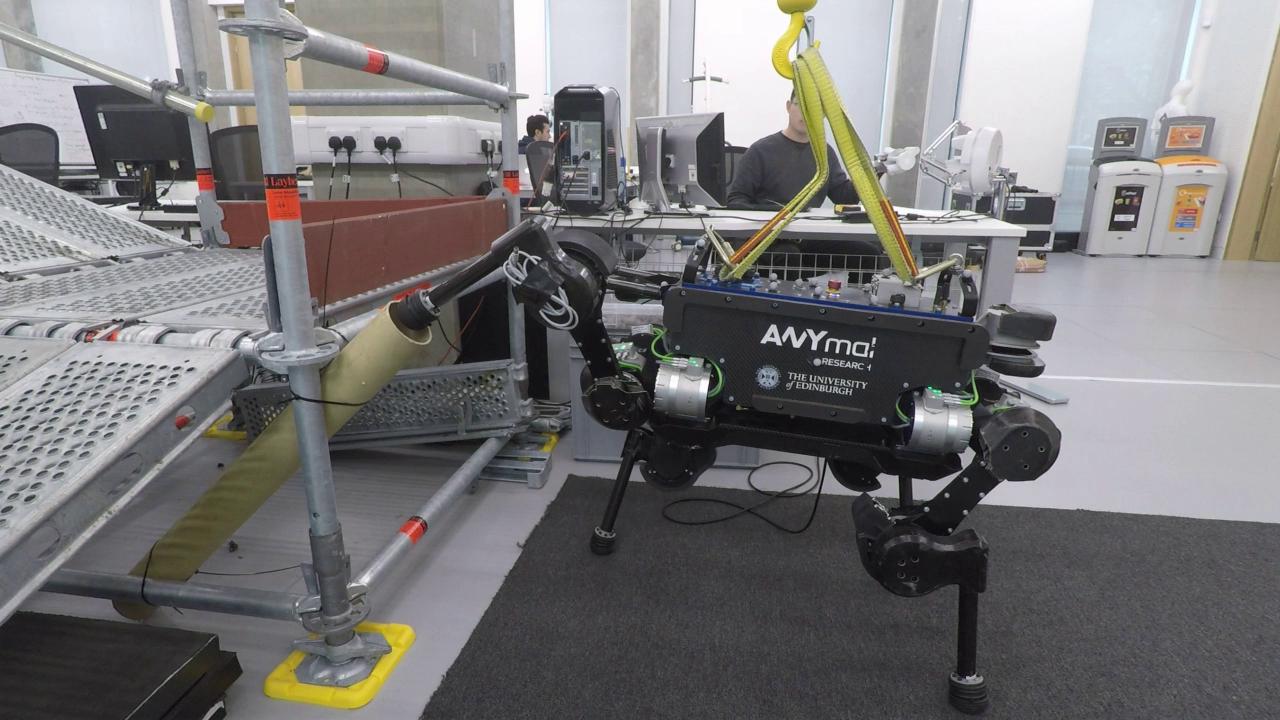}\put(90,50){\manuallabel{f:pipInsert3}{(c)}\ref{f:pipInsert3}}\end{overpic}~
\begin{overpic} [width=.48\linewidth]{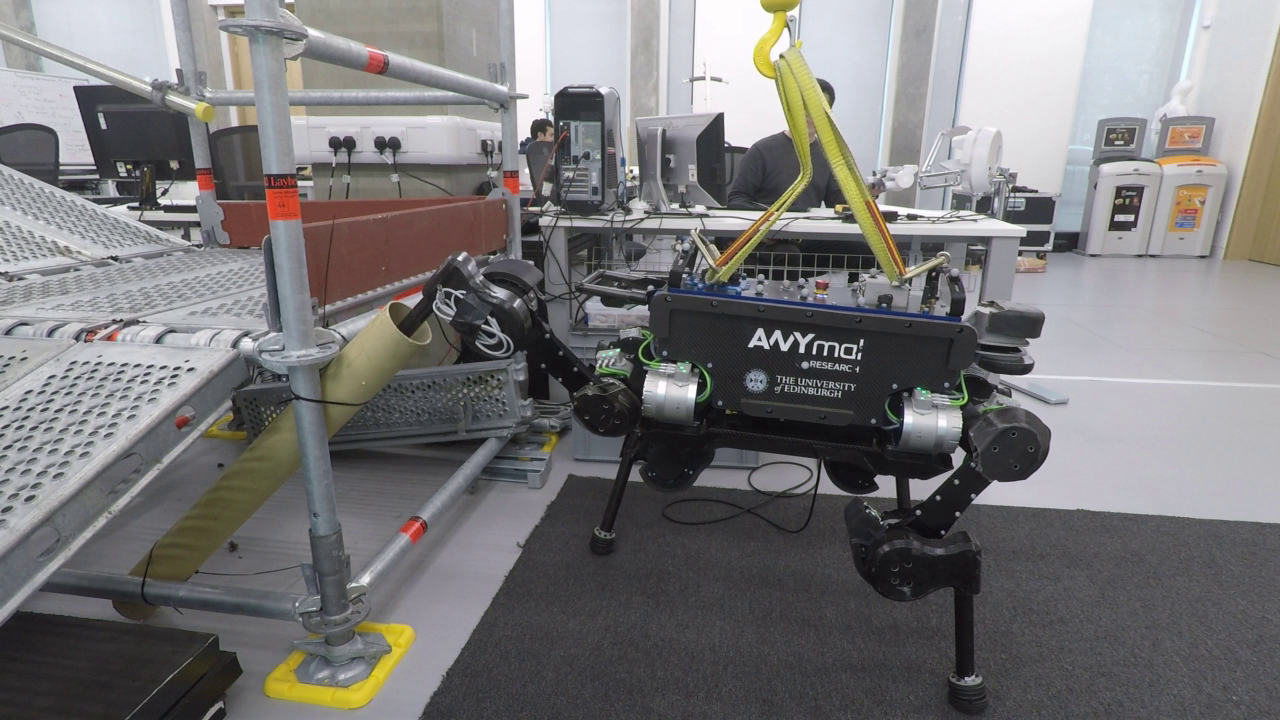}\put(90,50){\manuallabel{f:pipInsert4}{(d)}\ref{f:pipInsert4}} \end{overpic}
\\\vspace{-1mm} 
\caption{Teleoperating the foot into a pipe}  
\label{f:pip} 
\end{figure}

In this experiment we show-case the robot inserting its foot into a pipe, via teleoperation.
This scenario is relevant for industrial use and inspection tasks.
The aim of this experiment is to show the whole-body controller can control
foot posture, since inserting the foot into the pipe demands accurately
adjusting the foot orientation to be aligned with the pipe.
Fig. \ref{f:pip} shows the snapshots of the process.
We can see that the whole shank is finally inside the pipe.
Moreover the entrance of the pipe is quite high for ANYmal.
We rotated the second joint more than $^\pi/_2$ to reach the pipe.
During that we moved through a singular configuration as shown in the second snapshot.
The leg moved fast when trying to overcome singularity (as can be seen in the accompanying video), but this did not cause any issue on the robot, which shows the robustness of the our controller.

\subsection{Exploring surface}

\begin{figure}
\centering
\includegraphics [width=.48\linewidth]{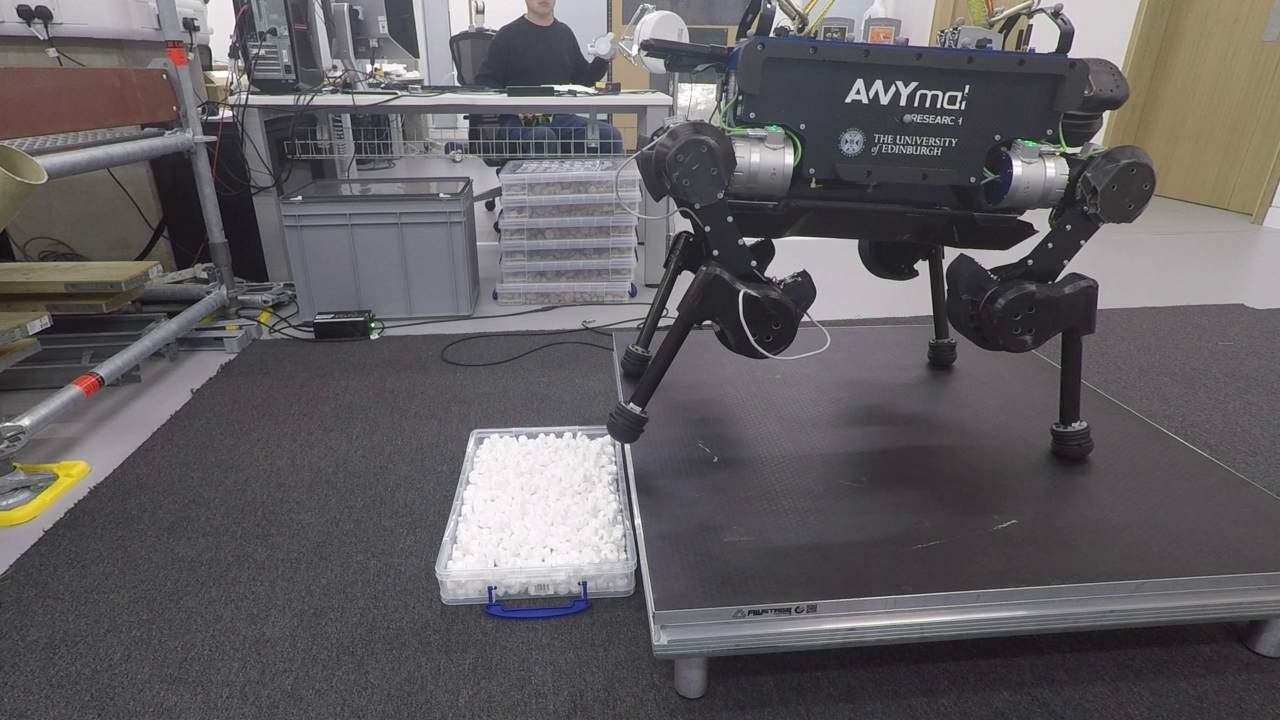}~
\includegraphics [width=.48\linewidth]{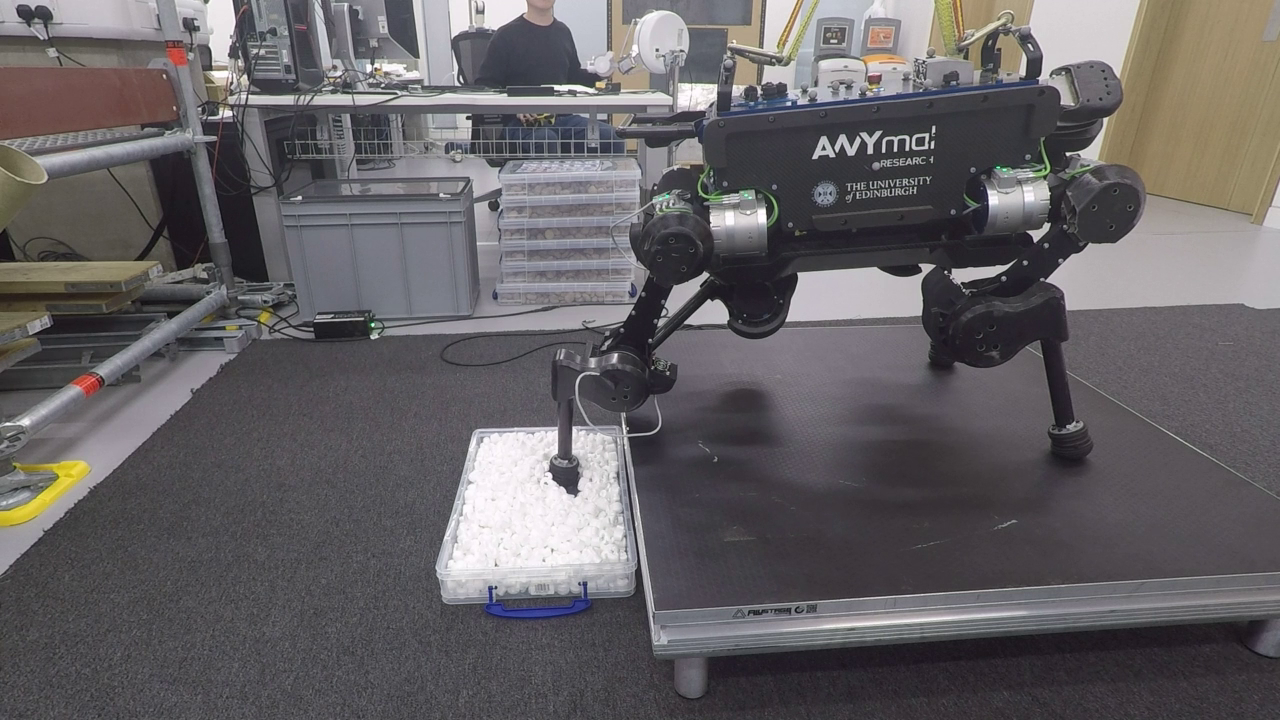}
\\\vspace{-1mm} 
\caption{Teleoperating the foot to reach the bottom covered by plastic peanuts and sliding on the surface to establish the depth of the reliable bottom surface.}  
\label{f:explore} 
\end{figure}

In real world deployment, visual perception of solid surfaces may be obscured by liquid, smoke, grass etc. The robot's locomotion planner needs to know the accurate elevation map to plan foot placement. In this experiment the operator teleoperates the robot to explore the surface covered by some light packing peanuts. We place the robot on a 20 cm high stage as shown in Fig. \ref{f:explore}, such that the robot needs to stretch its leg to reach the bottom. Benefiting from the hierarchical whole-body controller, the robot can automatically lower its torso to extend the reachability of the leg. The operator then explores the surface by sliding the foot back and forth. If we recorded the foot trajectory during sliding on the surface, we can establish the geometrical plane to update the elevation map. 

\subsection{Feasible force polytope boundary}

\begin{figure}[t!]
\centering
\begin{overpic} [width=.48\linewidth]{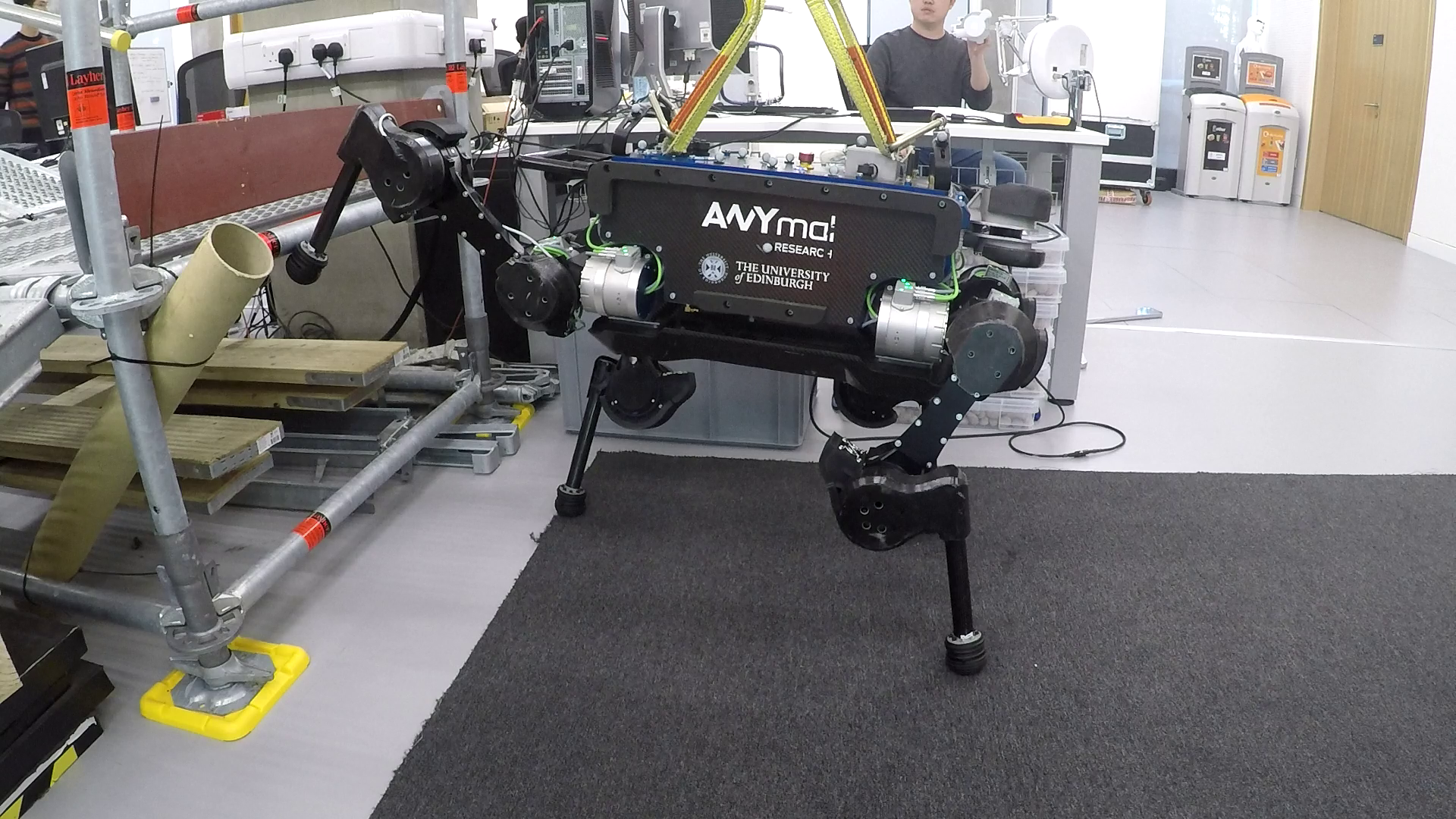}{\manuallabel{f:exp-wout-fx1}{(a)}\ref{f:exp-wout-fx1}}\end{overpic}~
\begin{overpic} [width=.48\linewidth]{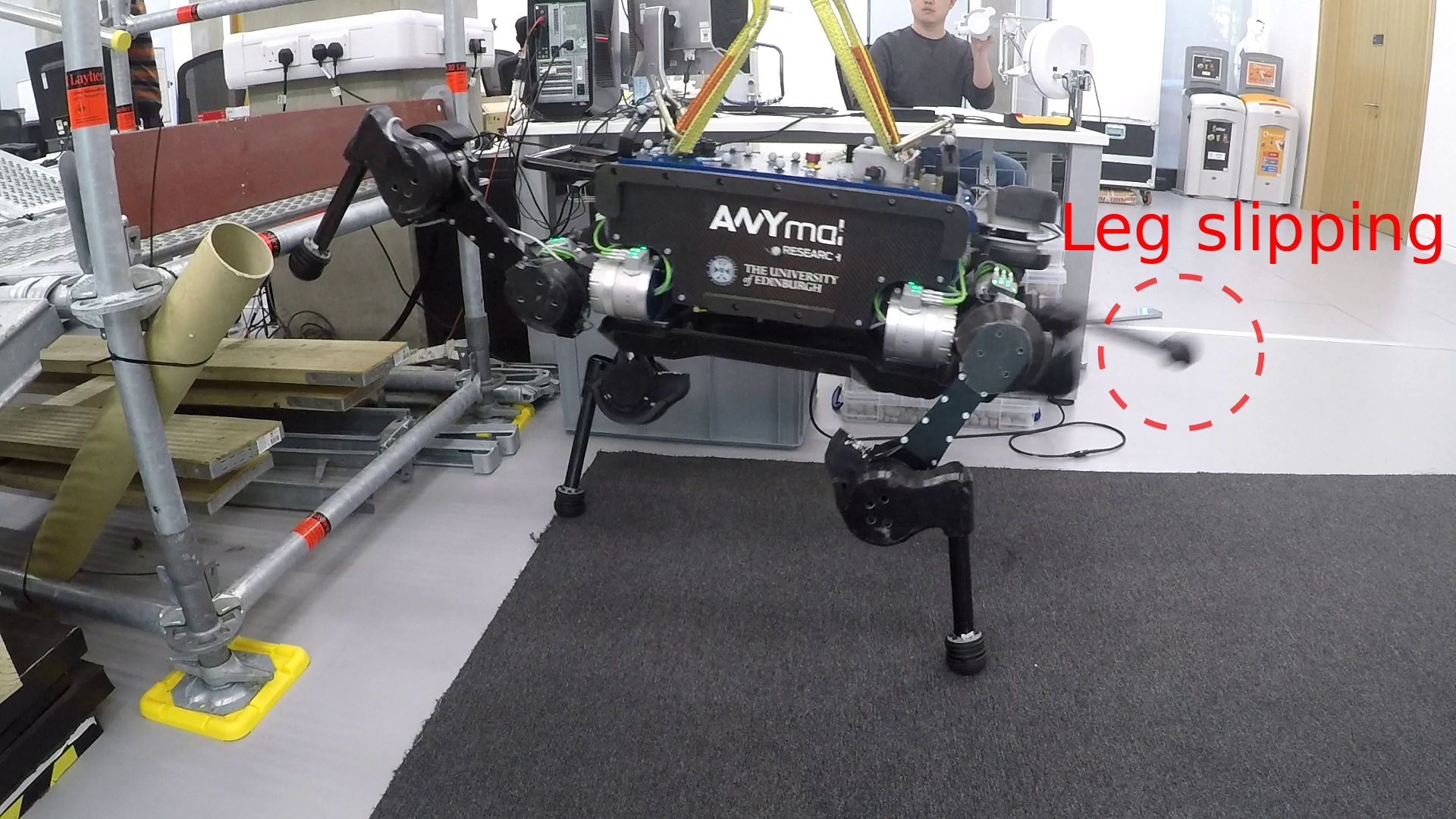}{\manuallabel{f:exp-with-fx1}{(b)}\ref{f:exp-with-fx1}} \end{overpic}
\caption{Pushing against scaffold. \ref{f:exp-wout-fx1} Starting to push, \ref{f:exp-with-fx1} Right hind leg slipping due to too much pushing force.}  
\label{f:pushing} 
\end{figure}

\begin{figure}[t!]
\centering
\begin{overpic} [width=.48\linewidth]{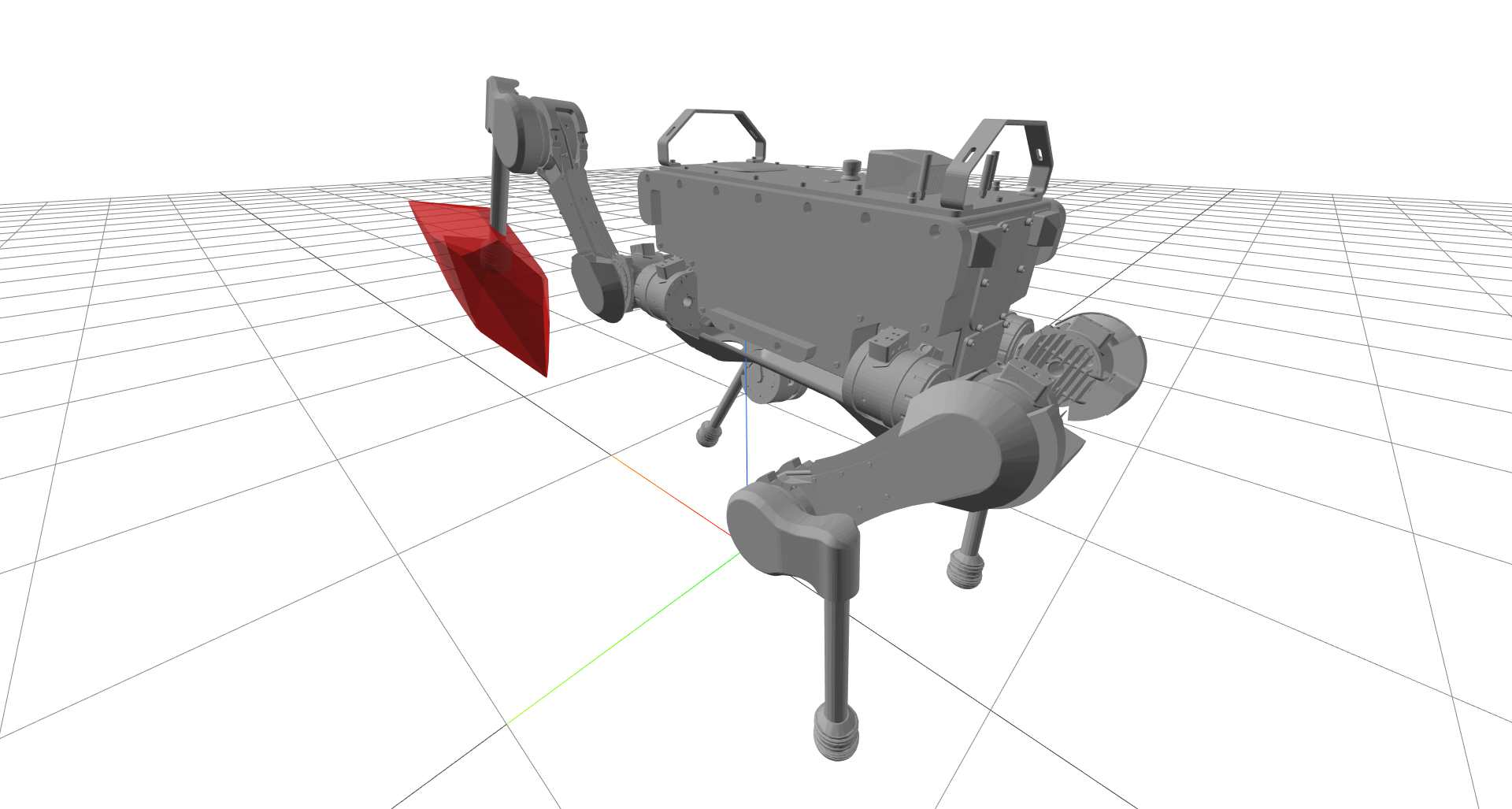}{\manuallabel{f:exp-wout-fx2}{(a)}\ref{f:exp-wout-fx2}}\end{overpic}~
\begin{overpic} [width=.48\linewidth]{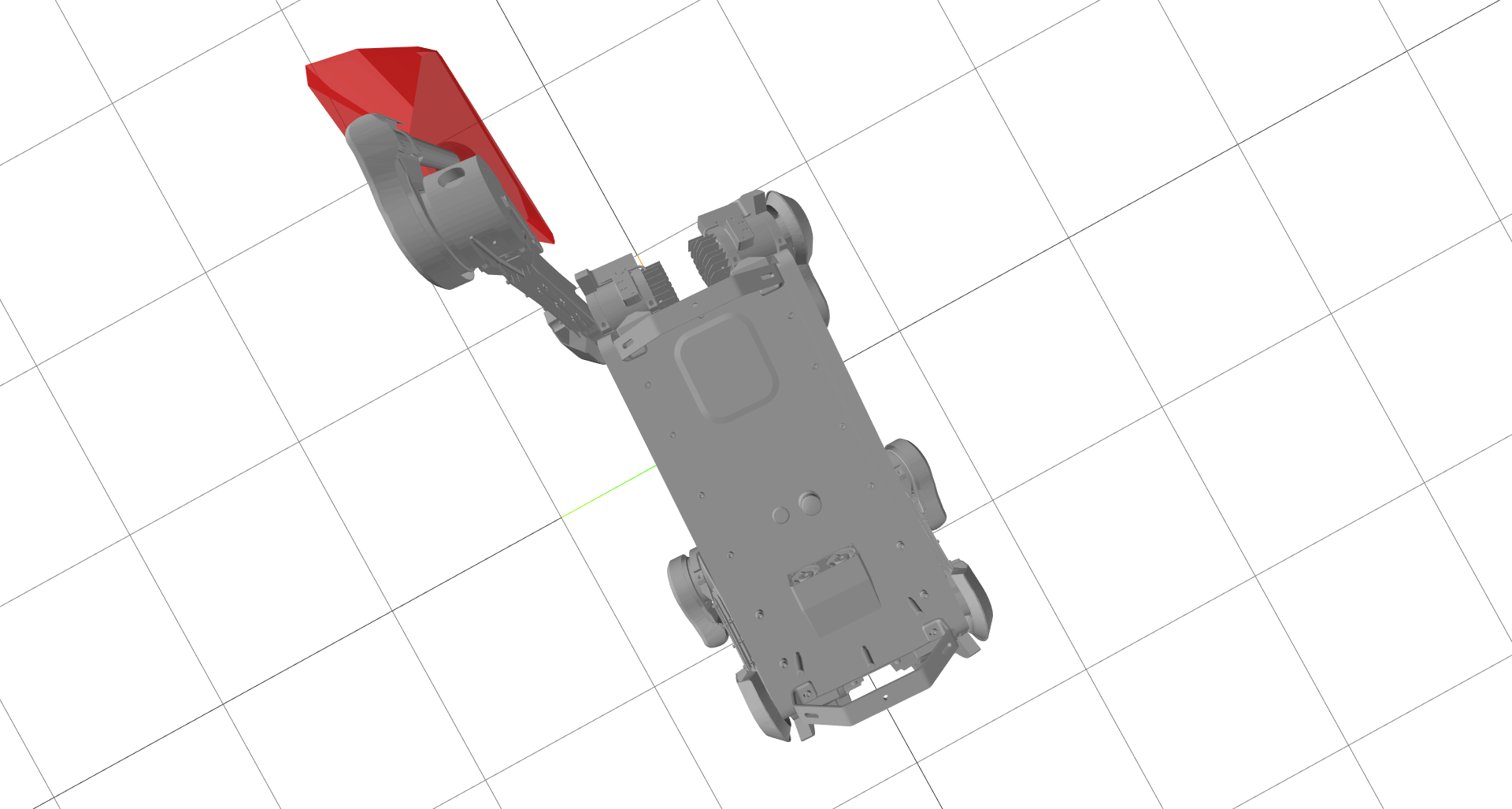}{\manuallabel{f:exp-with-fx2}{(b)}\ref{f:exp-with-fx2}} \end{overpic}
\caption{$\mathit{FFP}$ of the configuration in the pushing experiment. \ref{f:exp-wout-fx2} Side view, \ref{f:exp-with-fx2} Top view.}  
\label{f:pushingPolytope} 
\end{figure}

\begin{figure}
    \centering
    \includegraphics [width=.95\linewidth]{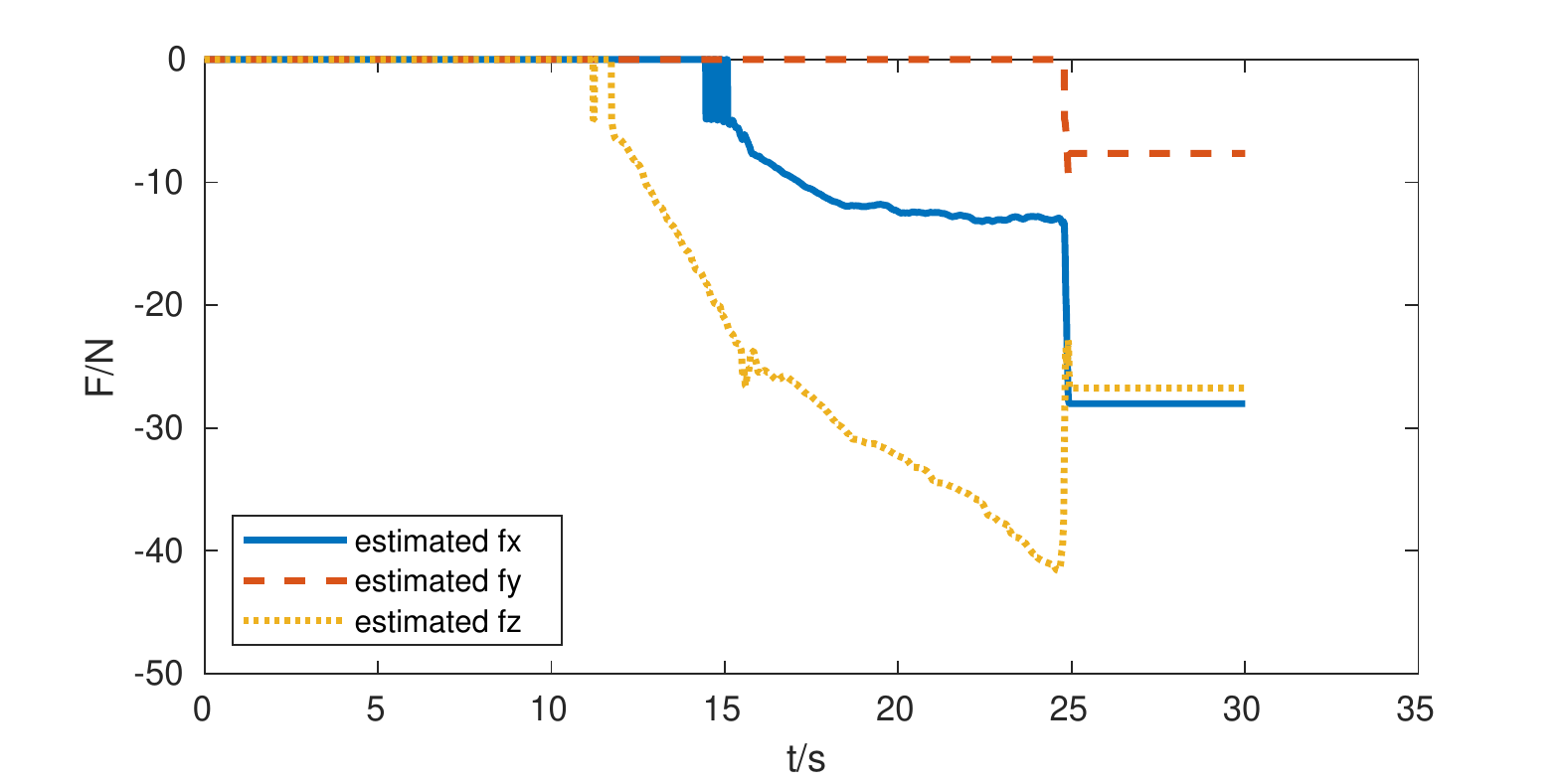}
    \\ \vspace{-1mm} 
    \caption{Estimated contact forces during pushing against a scaffold structure. The right hind foot started to slip at around \SI{25}{\second} because the contact forces was greater than the limit.}
    \label{f:forcePolytope}
\end{figure}

\begin{figure}
    \centering
    \includegraphics [width=.95\linewidth]{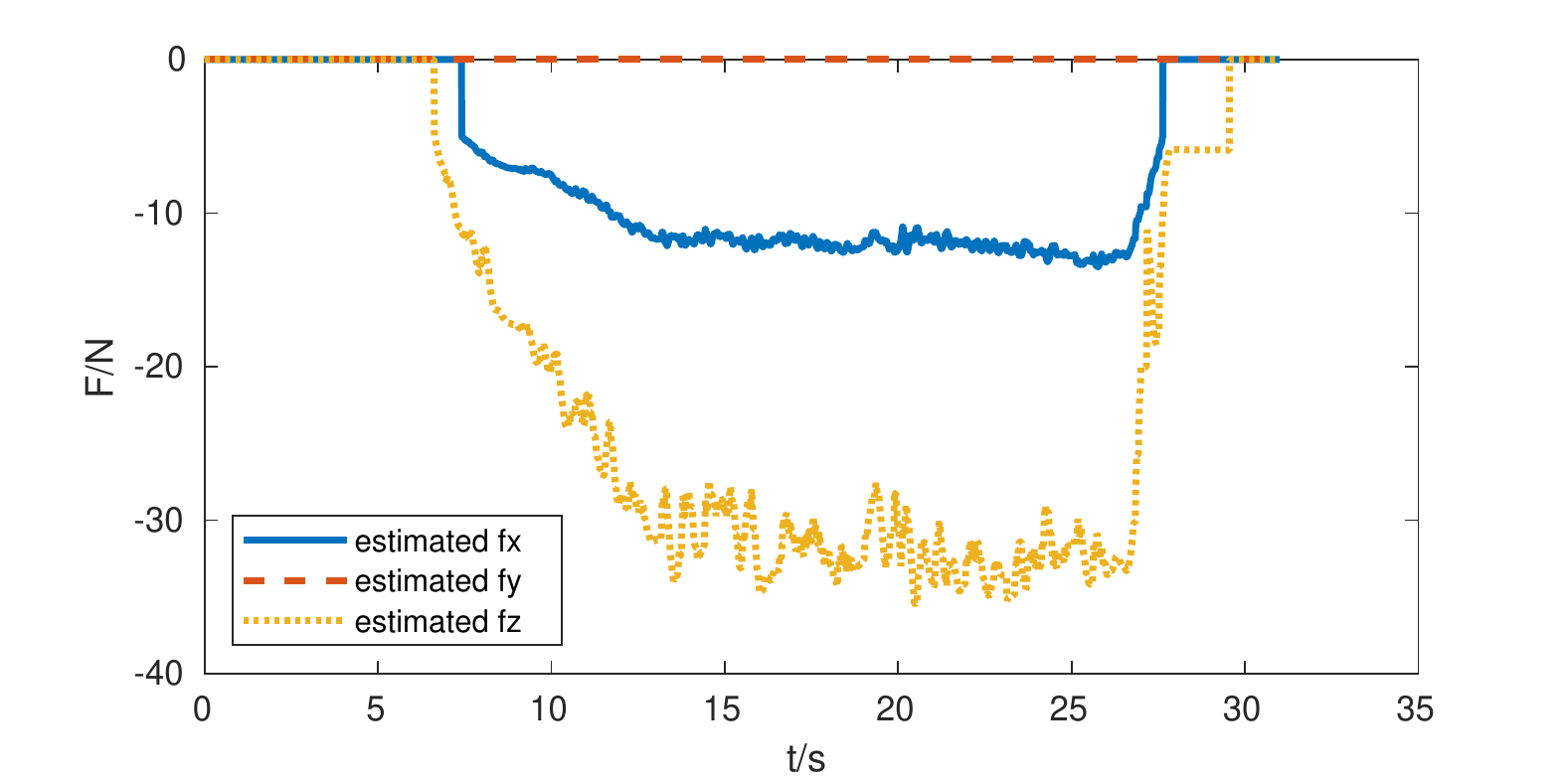}
    \\ \vspace{-1mm} 
    \caption{Estimated contact forces during pushing against a scaffold structure with $\mathit{FFP}$ boundary running. The oscillation caused by predefined pushing back forces around the boundary implies the boundary is triggered.}
    \label{f:vibration}
\end{figure}

In this experiment w e verify that setting a boundary on the joystick to
restrict acting force is necessary. The robot was operated to push against a
scaffolding structure as shown in Fig. \ref{f:pushing}.
The $\mathit{FFP}$ is calculated based on the measured configuration when the robot is pushing as shown in Fig. \ref{f:pushingPolytope}. The polytope for this configuration has 33 surfaces. The smallest infeasible force in this configuration is \SI{25.6}{\newton} with direction $\begin{bmatrix}0.08 &-0.99 & -0.11\end{bmatrix}^\top$. To verify the correctness of $\mathit{FFP}$, the operator kept pushing upward and forward in order to trigger the $\mathit{FFP}$ limitation without a boundary being set on joystick. The recorded contact forces are shown in Fig. \ref{f:forcePolytope}.
The right hind leg slipped quickly at around \SI{25}{\second} when the recorded forces jump up according to Fig. \ref{f:forcePolytope}. 
The recorded data after \SI{25}{\second} is irrelevant because the system crashed after detecting large slipping. 
We use the recorded forces to compute the projection along each direction of $\mathit{FFP}$. 
This shows that the projection of the contact force along with direction $\begin{bmatrix}-0.39 &-0.04 & 0.92\end{bmatrix}^\top$ is 30.69 at \SI{23.5}{\second}.
That is slightly greater than the model-predicted maximum feasible value 30.68 in that direction, 
implying that the computed $\mathit{FFP}$ is close to the real limit considering model error and force estimation error.
Subsequently, we repeated that experiment with enabled $\mathit{FFP}$ boundary on joystick. We shrunk the boundary to 80\% for safety. The oscillation of contact force shown in Fig. \ref{f:vibration} happened when the boundary was triggered. The increased stiffness of joystick prevented the operator from sending infeasible commands.

{}

{}

\section{CONCLUSIONS}

A hierarchical whole-body controller is successfully applied to foot posture teleoperation of a quadruped robot ANYmal. 
The output torques of the whole-body controller consists of two orthogonal components: 1) motion torques which are dedicated to execute mission commands without considering physical constraints, and 2) constraint torques that are only for satisfying physical constraints without generating any movement. The motion torques are computed by two analytical Cartesian impedance controllers decoupled by a dynamically consistent null-space projector. Motion control is performed by an analytical controller, such that only one QP optimization needs to be solved to cope with the inequality constraints. Two impedance controllers for the foot posture control and the torso control potentially overlap, but the null-space projector imposes a strict priority, which benefits the foot reachability. The foot impedance controller is also used to estimate contact forces when the manipulation foot is in contact with the environment. Experimental results show the estimation accuracy of this method for haptic teleoperation. In order to avoid falling over due to infeasible commands sent by operator, the CoM and the acting forces are effectively bounded by limiting the joystick movement. These boundaries not only protect the robot, they also guide the operator to move the joystick in feasible directions. Future work will aim to provide predictive online computing algorithms for the CoM boundary. 

\addtolength{\textheight}{-12cm}   




\section*{ACKNOWLEDGMENT}

This work has been supported by the following grants: EPSRC UK RAI Hub NCNR (EPR02572X/1) and ORCA (EP/R026173/1), THING project in the EU Horizon 2020 (ICT-2017-1), and by grant EP/L016834/1 for the University of Edinburgh RAS CDT from EPSRC.


\bibliographystyle{IEEEtran}
\bibliography{bibliography}

\end{document}